\documentclass{article}

\usepackage{times}
\usepackage{soul}
\usepackage{url}
\usepackage[hidelinks]{hyperref}
\usepackage[utf8]{inputenc}
\usepackage[small]{caption}
\usepackage{graphicx}
\usepackage{amsmath}
\usepackage{amsthm}
\usepackage{booktabs}
\usepackage{algorithm}
\usepackage[switch]{lineno}
\usepackage{comment}
\usepackage{algpseudocode}
\usepackage{amssymb}
\usepackage{subfig}

\title{Zero-shot Sim2Real Adaptation Across Environments}

\author{
Buddhika Laknath Semage*, Thommen George Karimpanal, Santu Rana,\\ Svetha Venkatesh \\
Applied Artificial Intelligence Institute\\
Deakin University\\
Geelong, Australia\\
*Email: bsemage@deakin.edu.au
}

\begin{document}

\maketitle

\begin{abstract}
Simulation based learning often provides a cost-efficient recourse
to reinforcement learning applications in robotics. However, simulators
are generally incapable of accurately replicating real-world dynamics,
and thus bridging the sim2real gap is an important problem in simulation based
learning. Current solutions to bridge the sim2real gap involve hybrid
simulators that are augmented with neural residual models.
Unfortunately, they require a separate residual model for each individual
environment configuration (i.e., a fixed setting of environment variables
such as mass, friction etc.), and thus are not transferable to new
environments quickly. To address this issue, we propose a Reverse Action
Transformation (RAT) policy which learns to imitate simulated policies
in the real-world. Once learnt from a single environment, RAT can
then be deployed on top of a Universal Policy Network to achieve zero-shot
adaptation to new environments. We empirically evaluate our approach
in a set of continuous control tasks and observe its advantage as
a few-shot and zero-shot learner over competing baselines.

\end{abstract}

\section{Introduction}
In recent years, deep reinforcement learning (RL) has been successfully
used to solve a number of complex physics related problems such as
solving a Rubik's Cube with a robotic hand \cite{akkaya2019solving},
manoeuvring objects \cite{DBLP:conf/rss/LiLH18}, etc. To circumvent
the sample inefficiency of deep learning methods used in these applications,
many of them use numerical physics-based simulators to generate cheaper,
synthetic experiences. Simulation-based learning methods
such as Universal Policy Networks (UPNs)\cite{up-net} additionally offer the benefit
of learning a range of policies across different environmental configurations
(e.g., mass, friction, etc.,) at one go, enabling quick adaptability
if the environment parameters change. However, the exact replication
of real-world dynamics in simulators is challenging due to the existence
of environment-specific transient dynamic factors (e.g., air-resistance),
which can be either exceptionally hard or too computationally intensive
to be modelled. Due to such discrepancies between the simulator and
real-world dynamics (i.e., reality-gap), policies learnt in simulators
generally do not perform well when they are directly transferred
to the real-world.

To bridge this reality-gap, hybrid simulators consisting of both numerical
as well as neural components have been studied by \cite{Ajay2018-ra,Ajay2019-vl,Johannink2019-bp,Ba2019-pz}.
While these approaches have been shown to bridge the reality-gap,
they are designed to be trained separately for each individual real-world environment,
and have to be retrained when the environment parameters change.
\emph{Action Transformation (AT)} policy by \cite{hanna2021grounded,stocgat20,karnan2020reinforced}
learns to transform actions in the simulator to mimic the state transitions 
in the real world, thereby, making the learnt policy applicable to the real world.
However, they are not compatible with pretrained policies such as UPNs because task-policies
need to be learnt on the transformed action set, which limits the adaptability of the agent. Thus,
learning to bridge the reality-gap for a range of environmental settings
without retraining is still an open problem.

\begin{figure}
\begin{centering}
\includegraphics[width=0.9\columnwidth]{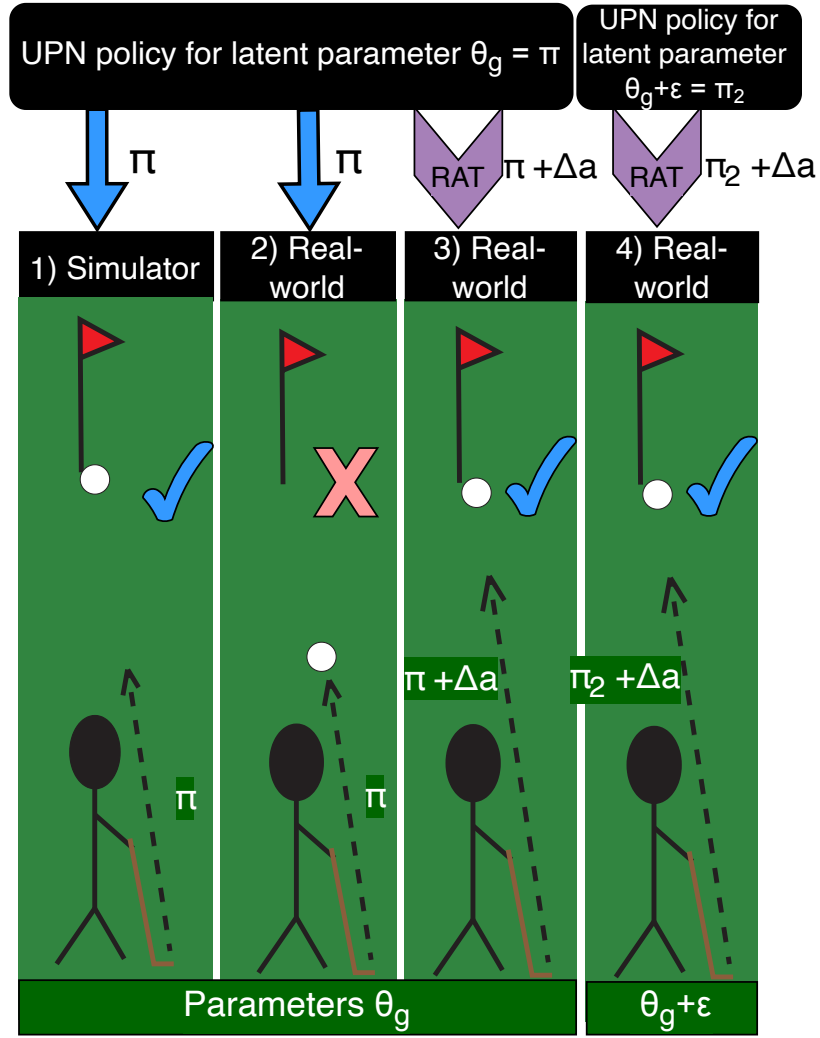}
\par\end{centering}
\caption[Combining \emph{RAT} with a UPN for fast adaptability]{\label{fig:RAT-intro}Combining \emph{RAT} with a UPN for fast adaptability
of pretrained simulated policies to real-world conditions. For a task
of golf played under known modellable latent parameters $\theta_{g}$
(e.g., friction), (1) the optimal simulated policy $\pi$ from UPN
for parameters $\theta_{g}$ achieves the task in the simulator. (2)
When the same policy $\pi$ is transferred to the real-world, due
to the reality gap (e.g., rolling friction in the real-world) the
simulated policy does not achieve the task. (3) \emph{RAT }transforms
policy $\pi$ such that the resulting trajectory, when rolled out
in the real-world, matches the simulated trajectory through a $\Delta a$
correction in the action. (4) For an adjacent environment 
with latent parameters $\theta_{g}+\epsilon$, 
the respective UPN policy $\pi_{2}$ is transformed
by \emph{RAT }using the same $\Delta a$ correction. }
\end{figure}

To address these shortcomings, we propose a \emph{Reverse Action Transformation
(RAT)} policy that combines the adaptability of UPN with the concept
of action transformation. Figure \ref{fig:RAT-intro} shows how a
golf-playing robot can bridge the sim-to-real gap using corrective
actions from the \emph{RAT} policy. Instead of learning to reproduce real-world
dynamics in the simulator as done in \emph{GAT} \cite{gat2017},
in \emph{RAT}, we aim to learn the reverse transformation --- making
real-world trajectories mimic the optimal trajectories learnt in the
simulator. If the reality-gap remains fixed locally across adjacent
environment configurations, \emph{RAT} policies learnt in one environment
will also transfer to those adjacent environments, making the combination
of UPN+\emph{RAT} a vehicle for adaptable reality-gap-free policy
learning. We add that our use of 'reality-gap-free' learning utilises the predictability
of the agent, as the agent will see the real-world as a simulator
and thus would be able to take predictable optimal actions as per
its simulation-learnt policy. Even though such a policy may be slightly
suboptimal (e.g., if the reality-gap helps to discover a better policy), 
it makes our learning robust to test time changes in the environment configuration.
Specifically, we first train a UPN on a relatively small number of
simulator parameters to learn optimal simulated policies (step 1,
Fig. \ref{fig:RAT-intro}), which may not transfer well to the real-world
(step 2, Fig. \ref{fig:RAT-intro}). Next, we train \emph{RAT} by
sampling different reality-gap transitions in the simulator to learn
a robust initial policy. Initialised with this policy, we then deploy the combination
of UPN+\emph{RAT} policy in the real-world to bridge the reality-gap
(step 3, Fig. \ref{fig:RAT-intro}). Subsequently, when an environment
change is detected (e.g., change in ground friction due to rain),
we appropriately set the parameter of the UPN for the new environment
and use the UPN+\emph{RAT} without any retraining to achieve zero-shot
adaptation (Step 4, Fig. \ref{fig:RAT-intro}). One of the key intuitions
behind learning \emph{RAT} is that in the presence of moderate reality
gaps, it would be relatively more sample efficient to learn an explicit
mapping from simulated policies to their ideal real-world counterparts,
as opposed to adapting each task-specific simulation policy to the real-world.
Experiments on six continuous control tasks show that our approach
outperforms other UPN-based adaptable baselines. 

In summary, the main contributions of our study are:
\begin{enumerate}
\item Proposing \emph{Reverse Action Transformation (RAT), }an adaptable policy that
can correct real-world policies to follow optimal simulated trajectories
learnt from a Universal Policy Network.
\item Formulating and implementing \emph{RAT} policy with a robust initial policy to improve the sample efficiency of learning.
\item Empirically demonstrating improved real-world transfer on a set of continuous control tasks.
\end{enumerate}

\section{Background}
Simulators are a cheap source of synthetic data to alleviate the sample
inefficiency of deep learning methods \cite{mnih2015human} that
are being used with many RL applications today. However, for successful
sim2real transfer, simulation parameters (e.g., friction, restitution)
often need to be tuned to match the real-world dynamics (i.e., system
identification or grounding). A relatively simple basis for this is
to simply minimise the differences between the simulation and real-world
trajectories \cite{using-inaccurate-models,10.5555/3304889.3305112,DBLP:conf/icra/ChebotarHMMIRF19,du2021auto,allevato2020tunenet,Allevato:2020ui}.
Direct policy search is another approach for system identification,
where optimal simulation parameters are discovered by evaluating simulated
policies in the real-world to find the parameters corresponding to
the highest performing simulated policy \cite{farchy2013humanoid,up-net,yu2018policy,yu2019sim,evalutionary-direct-search}. 

Many direct policy search methods, although imperfect at modelling
the real-world, carry the benefit of adaptability due to their maintenance
of many pretrained simulation policies which are directly transferable
to the real-world after the grounding process. In this light, Universal
Policy Networks (UPN) \cite{up-net} has become a useful tool, as
it offers a framework for learning policies for a continuous range
of simulation parameters by training only on a limited set of parameters.
With such a trained UPN, neural networks \cite{up-net}, Bayesian
Optimisation \cite{yu2019sim} and evolutionary algorithms \cite{yu2018policy}
can be used to efficiently search for an optimal set of parameters
that yield the best policy to be transferred to the real-world. For
the purpose of this study, we assume the knowledge of real-world parameter
values and thus, obviate the grounding process from both our method
and the baselines.

In addition to grounding, hybrid simulators examine combining numerical
simulators with neural components to reduce the mismatch in the dynamics
of the simulator and the real-world \cite{Ajay2018-ra,Ajay2019-vl,Ba2019-pz,pmlr-v87-golemo18a,DBLP:journals/corr/abs-2101-06005}.
Although they have been shown to improve modelling capabilities compared
to analytical simulators, since the focus is to solely improve the
dynamics mismatch, policies learnt on such hybrid simulators would
need to be re-trained when the residual layer, which models factors
such as air resistance or rolling friction, changes. Given that such
factors are usually highly transient, frequent changes to the residual
layer would be needed. As such, policies trained on hybrid simulators
are ill-equipped for learning adaptable policies under changing environment
conditions. Residual policies \cite{Silver2018-jq,TossingBot,Johannink2019-bp}
follow a similar concept, where a classical controller is augmented
with a correction policy in the real-world using the task reward.
Since they learn a policy corresponding to a single parameter setting,
this solution is also not adaptable when the environment conditions
change.

Instead of trying to replicate real-world dynamics through the transition
function, \emph{Grounded Action transformation (GAT) }\cite{gat2017}
and other related approaches \cite{stocgat20,karnan2020reinforced,DBLP:conf/rss/ZhangZLL21,DBLP:conf/nips/DesaiDKWHS20}
directly modified the simulated policy to match the dynamics of the
real-world. In these works, the simulator is grounded to follow the
real-world using the learnt action mappings. While our study was inspired
from this line of research, we try to address several drawbacks with
\emph{GAT: }1) These works are incompatible with pretrained policies (e.g., universal policies),
because the simulated task policies need to be learnt on transformed actions,
2) Since the simulator's state space can be more limited than that of the
real-world, trying to imitate the real-world using such a simulator
can lead to subpar results, 3) deploying these methods on unknown environments with 
a different reality-gap will require the action 
mappings to be updated, which in turn will require 
re-learning of the task policies. In contrast to this, we enable 
the same policies to be retained, requiring adaptation 
of just the \emph{RAT} policy, which is significantly easier to train. 

Learning simulated policies robust to real-world noise is another
approach related to our work. Domain randomisation (DR) \cite{8202133,BayesSimRamosPF19,data-driven-dr,drboMuratoreEGP21,mozian2020learning,DBLP:conf/rss/SadeghiL17,matas2018sim,akkaya2019solving,DBLP:conf/iclr/RajeswaranGRL17}
has become a common approach, where a robust policy is learnt by training
on a range of parameters. While it provides a relatively convenient
approach for sim2real transfer without specifically grounding the
simulator, the performance of a DR policy heavily relies on the parameter
distribution on which it was trained. Particularly, when the parameter
range is large, DR policies have been known to produce extremely conservative
policies\cite{data-driven-dr} compared to regular policies.

\section{Method}
In this study, we propose a mechanism to bridge the reality-gap by
learning a policy to correct a pretrained simulation policy, such
that the corrected policy produces real-world trajectories that closely
match the optimal trajectory learnt in the simulator. The simulated
environment $\psi^{sim}$ is parameterised by modellable latent parameters
$\theta$ for which we can model accurate simulation dynamics using
standard numerical simulators without extensive computational costs.
The real-world $\psi$ is parameterised by $\phi=[\theta,U]$
($\theta,U\in\mathbb{R}^{d}$), where $U$ represents
unmodellable parameters that are difficult to be modelled universally (e.g.,
air-resistance, rolling friction). Learning tasks in the simulator
and the real-world are represented as MDPs $\mathcal{M}_{sim}=\:<\mathcal{S},\mathcal{A},\mathcal{T}_{sim},\theta,\mathcal{R}>$
and $\mathcal{M=}<\mathcal{S},\mathcal{A},\mathcal{T},\phi,\mathcal{R}>$
respectively, where $\mathcal{S}$ is the state space, $\mathcal{A}$ is the action space,
$\mathcal{T}_{sim}$ and $\mathcal{T}$ are respectively the simulation and real-world
transition functions with $\mathcal{S}\times \mathcal{A}\rightarrow \mathcal{S}$ mappings 
and $\mathcal{R}:\mathcal{S}\times \mathcal{A}\rightarrow\mathbb{R}$
is the reward function. 

Under this formulation, we utilise an RL agent design commonly known
as Universal Policy Network (UPN) to learn in simulation, a large
range of policies corresponding to different latent parameter configurations.
A UPN is learned on the MDP $\mathcal{M}_{sim}$ by sampling
parameters $\theta$ in the simulation and learning a policy network
capable of providing policies conditioned on latent parameters $\theta$.
To achieve this goal, UPN's state is constructed by appending the
task's observable state (e.g., object positions, velocities) with
the sampled $\theta$ (e.g., friction, restitution) values in a given
range. The UPN is then trained in simulation with RL to learn policies
over the range of the provided latent parameter values, while also
generalising to unseen latent parameters in the range. 

\begin{figure}
\begin{centering}
\includegraphics[width=0.95\columnwidth]{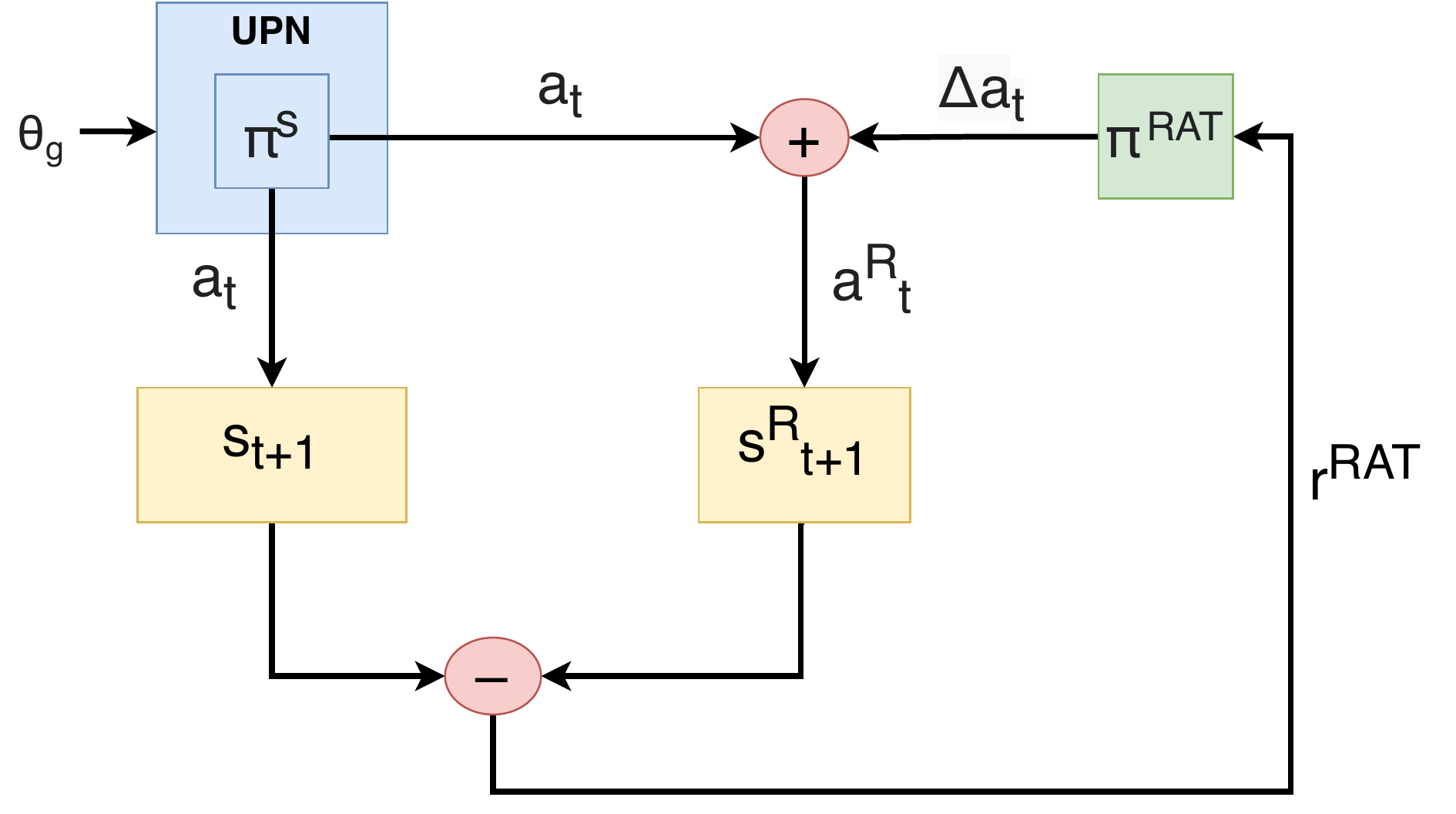}
\par\end{centering}
\caption[The process of learning Reverse Action Transformation (RAT).]{\label{fig:rat-learning} The process of learning Reverse Action
Transformation (\emph{RAT}). The \emph{RAT }policy $\pi^{RAT}$ outputs
$\Delta a_{t}$ which is the amount of adjustment needed on the UPN
$\pi^{S}$ action $a_{t}$ (conditioned by latent parameter
$\theta_{g}$) to follow the simulated trajectory in the real-world.
The adjusted action $a_{t}^{R}$ is executed in the real-world which
results in the state $s_{t+1}^{R}.$ The difference between the simulator's
$s_{t+1}$ state after executing $a_{t}$ and the real-world's $s_{t+1}^{R}$
is used as a negative reward $r^{RAT}$ in learning $\pi^{RAT}$. }
\end{figure}

In this workflow, we first pretrain a UPN, and aim to appropriately
correct a suitable policy drawn from it, so that the resulting policy
produces desirable trajectories in a previously unseen real-world.
For this purpose, we learn a \emph{Reverse Action Transformation (RAT)
} policy that appropriately corrects the simulated policy to reproduce
the corresponding desired simulated trajectory in the real-world (Fig.
\ref{fig:rat-learning}). \emph{RAT} would then transfer to span other
environments (with slightly different latent parameters) as long as
the reality-gap remains the same.

\subsection{\emph{Reverse Action Transformation (RAT)} Policy\label{subsec:reverse-action-transformation}}

One of the key intuitions behind learning \emph{RAT} is that in the
presence of large reality gaps, it would be relatively more sample
efficient to learn an explicit mapping from simulated policies to
their ideal real-world counterparts, as opposed to adapting each task-specific
simulation policy to the real-world. For a measured ground truth modellable
latent parameter $\theta_{g}\in\theta$ and an unknown reality-gap
$\mu_{g}\in U$, we learn a correction policy $\pi_{\theta_{g},\mu_{g}}^{RAT}$
in a real-world $\psi_{\phi_{g}}$ (parameterised by $\phi_{g}=[\theta_{g},\mu_{g}]$)
which when employed in conjunction with the UPN, imitates the UPN's
$\theta_{g}$ conditioned policy $\pi_{\theta_{g}}^{S}$ learnt in
a simulated environment $\psi_{\theta_{g}}^{sim}$. This approach
is similar to the idea of \emph{Action Transformations }(\emph{AT})
\cite{hanna2021grounded,stocgat20,karnan2020reinforced}, but instead
of learning a simulated policy to reproduce the real-world trajectories,
we learn the reverse (hence the name\emph{ Reverse Action Transform}),
by learning corrections to the trained simulation policy $\pi_{\theta_{g}}^{S}$
in order to reproduce the corresponding simulated trajectories in
the real-world (Fig. \ref{fig:rat-learning}). %

To closely emulate simulated trajectories in the real-world through
\emph{RAT}, we assume the existence of such a real-world policy. %
Secondly, for the purpose of this study, we assume ground truth latent parameters
of the original environment and environments corresponding to adjacent
latent parameter configurations to be known, perhaps through direct
measurements and/or policy search based grounding methods \cite{farchy2013humanoid,10.5555/3304889.3305112}.

We learn the \emph{RAT} policy through RL by defining an additional
MDP framework: $\mathcal{M^{RAT}}=$ $\mathcal{<S^{RAT}},\mathcal{A^{RAT}},\mathcal{T^{RAT}},\mathcal{R^{RAT}},\theta,U>$
where $\mathcal{S^{RAT}}$ is the \emph{RAT} state space, $\mathcal{A^{RAT}}$
is the \emph{RAT} action space, $\mathcal{T^{RAT}}:\mathcal{S^{RAT}}\times\mathcal{A^{RAT}}\rightarrow\mathcal{S^{RAT}}$
is the \emph{RAT} transition function and $\mathcal{R^{RAT}}:\mathcal{S^{RAT}}\times\mathcal{A^{RAT}}\rightarrow\mathbb{R}$
is the reward function used to train the \emph{RAT} policy. The state
$s_{t}^{RAT}\in\mathcal{S^{RAT}}$ at a given timestep $t$ is constructed
by concatenating the real-world state $s_{t}^{R}\mathcal{\in S}$
with the greedy action of the trained simulation policy obtained from
the UPN.

\[
s_{t}^{RAT}=[s_{t}^{R},\thinspace\pi_{\theta_{g}}^{S}(s_{t})]
\]

Instead of directly imitating the simulated policy $\pi_{\theta_{g}}^{S}$,
\emph{RAT} follows an approach inspired by \cite{karnan2020reinforced},
by predicting the difference of actions $\Delta a_{t}\in\mathcal{A^{RAT}}$
between the simulated and the real-worlds. \cite{karnan2020reinforced}
suggested that such an approach improved learning due to the normalising
effect on the action output space of the corresponding neural network.

As explained later in Eq. \ref{eq:rat-reward}, the reward function
for training \emph{RAT} is determined by measuring the distance between
the simulated and real-world states when the corresponding actions
are taken. While the simulated action is determined by $\pi_{\theta_{g}}^{S}$,
the real-world action $a_{t}^{R}$ is determined by adjusting the
action from the simulated policy $\pi_{\theta_{g}}^{S}$ by the corresponding
\emph{RAT} action $\Delta a_{t}$ for that \emph{RAT} state $s_{t}^{RAT}$:

\begin{equation}
\begin{aligned}\Delta a_{t} & \gets\pi_{\theta_{g},\mu_{g}}^{RAT}(s_{t}^{RAT})\\
a_{t}^{R} & =\pi_{\theta_{g}}^{S}(s_{t})+\Delta a_{t}
\end{aligned}
\label{eq:rat-correction}
\end{equation}

We execute the action $a_{t}^{R}$ in the real-world and the UPN policy
$\pi_{\theta_{g}}^{S}(s_{t})$ in the simulated world to transit to
the states in next timestep $t+1$. Then, the squared Euclidean distance
between the real-world's state $s_{t+1}^{R}$ and simulated environment's
state $s_{t+1}$ in timestep $t+1$ is computed as a negative reward
$r^{RAT}\in\mathcal{R^{RAT}}$ for training \emph{RAT}.

\begin{equation}
r^{RAT}=-|s_{t+1}^{R}-s_{t+1}|_{2}^{2}\label{eq:rat-reward}
\end{equation}

Following the above steps iteratively, the \emph{RAT} policy $\pi_{\theta_{g},\mu_{g}}^{RAT}$
learns to produce the appropriate correction $\Delta a_{t}$ to produce
simulation-like trajectories in the real-world. \emph{RAT} policy
training is detailed in Algorithm \ref{alg:rat-train}.

\begin{algorithm}[htp] 
\caption{$RAT$ Policy Training ($Train\_RAT\_Policy$)} 
\label{alg:rat-train} \hspace*{\algorithmicindent}
\textbf{Input:} \begin{algorithmic}[1]
\State $\theta_{g}$ - ground-truth modellable parameters
\State $\mu_{g}$ - ground-truth unmodellable parameters (unknown)
\State $\pi^{S}$ - a trained UPN for the task
\State $\psi_{\phi}$ - real-world, $\psi^{sim}_{\theta}$ - a parameterisable simulator
\State $reset$ - simulator reset at each step?

\State \textbf{Output:} Trained $RAT$ correction policy $\pi_{\theta_{g}, \mu_{g}}^{RAT}$\\

\If {a $RAT$ initial policy given}
\State Initialise $RAT$ policy $\pi_{\theta_{g},\mu_{g}}^{RAT} \gets $ $RAT$ initial policy
\Else
\State Initialise $RAT$ policy $\pi_{\theta_{g},\mu_{g}}^{RAT} \gets \varnothing$ 
\EndIf

\State $s_{t}^{R} \gets$ Initialise state from real-world task $\psi_{\phi}$
\State $s_{t} \gets s_{t}^{R}$, $\psi^{sim}_{\theta_{g}}$ state $\gets$ $s_{t}^{R}$

\For{Each step of the episode} 

\State //Get UPN action for the state $s_{t}$ and latent parameters $\theta_{g}$
\State $a_{t} \gets \pi_{\theta_{g}}^{S}(s_{t})$
\State $s_{t+1} \gets$ Execute $a_{t}$ in simulator $\psi^{sim}_{\theta_{g}}$

\State //Get $RAT$ corrected action
\State $a_{t}^{R} \gets a_{t} + \pi_{\theta_{g},\mu_{g}}^{RAT}(s_{t}^{RAT} = [s_{t}^{R}, a_{t}])$
\State $s_{t+1}^{R} \gets$ Execute $a_{t}^{R}$ in the real-world $\psi_{\phi}$

\State $\pi_{\theta_{g},\mu_{g}}^{RAT} \gets $ Use $s_{t+1}$, $s_{t+1}^{R}$ with Eq. \ref{eq:rat-reward} to calculate the reward and update $RAT$ policy using the selected RL algorithm

\State $s_{t}^{R} \gets s_{t+1}^{R}$

\If {$reset$}
\State //Change simulator's state to real-world state
\State $\psi^{sim}_{\theta_{g}}$ state $\gets$ $s_{t}^{R}$, $s_{t} \gets$ $s_{t}^{R}$
\Else
\State $s_{t} \gets s_{t+1}$
\EndIf

\EndFor

\end{algorithmic} 
\end{algorithm}

\subsection{Robust \emph{RAT} Initialisation\label{subsec:rat-prior}}

As the reality-gap in the real-world is unknown, we require the initial
\emph{RAT} policy to be robust to span different possible reality-gap values.
To incorporate this property, we use domain randomisation in the simulator
to learn a robust \emph{RAT} initial policy $\bar{\pi}_{\theta_{g}}^{RAT}$
by training it using a range of reality-gaps. For a defined $l$ lower
and $h$ upper limits, we uniformly sample a simulated reality gap
value $\bar{\mu}\sim Uniform(l,h)$, $\bar{\mu}\in U$ in
each episode, using which we model a hypothetical real-world $\psi_{\bar{\phi}}$
parameterised by $\bar{\phi}=[\theta_{g},\bar{\mu}]$. Using $\psi_{\bar{\phi}}$
and its transition function $\bar{\mathcal{T}}$, the subsequent hypothetical
real-world state $\bar{s}_{t+1}^{R}$ is given as. 

\begin{equation}
\bar{s}_{t+1}^{R}\gets\bar{\mathcal{T}}(s_{t},\bar{a}_{t}),\label{eq:rat-prior}
\end{equation}

Here, $\bar{a}$ is the corrected action obtained using robust \emph{RAT}
initial policy $\bar{\pi}_{\theta_{g}}^{RAT}$, 

\[
\bar{a}_{t}\gets a_{t}+\bar{\pi}_{\theta_{g}}^{RAT}(\bar{s_{t}}^{RAT}=[s_{t},a_{t}]),
\]

where $a_{t}$ is the UPN action corresponding to ground truth latent
parameters $\theta_{g}$, which produces state $s_{t+1}\gets\mathcal{T}_{sim}(s_{t},a_{t})$
from state $s_{t}$. The obtained states $\bar{s}_{t+1}^{R}$ and
$s_{t+1}$ are used to obtain the reward (as in Eq. \ref{eq:rat-reward})
for training the \emph{RAT} initial policy. 

The steps for training the \emph{RAT} initial policy are given in Algorithm
\ref{alg:rat-prior}. Once trained, the \emph{RAT} initial policy is used as
the initial estimate of the \emph{RAT} policy when used in a real-world
environment. Depending on the availability of a real-world interaction budget, the \emph{RAT}
policy then can be improved (i.e., adapted) from real-world interactions as an optional few-shot 
learning workflow (not explored in this study).

\begin{algorithm}[ht] 
\caption{Robust $RAT$ Initial Policy Training ($Train\_RAT\_Initial\_Policy$)} 
\label{alg:rat-prior} \hspace*{\algorithmicindent}
\textbf{Input:} \begin{algorithmic}[1]
\State $M$ - number of sim. episodes to run
\State $l$,$h$ - lower and upper bounds for reality-gap sampling
\State $\theta_{g}$ - ground-truth modellable parameters
\State $\psi^{sim}_{\theta}, \bar{\psi}^{sim}_{\bar{\phi}}$ - two parameterisable simulators

\State \textbf{Output:} Trained robust $RAT$ initial policy $\bar{\pi}_{\theta_{g}}^{RAT}$\\
\State Initialise $RAT\_initial$ policy $\bar{\pi}_{\theta_{g}}^{RAT} \gets \varnothing$ 

\For{$m \gets 1$ to $M$} 
\State $\bar{\mu} \sim Uniform(l, h)$ //sample a reality-gap value
\State $\bar{\phi}=[\theta_{g},\bar{\mu}]$
\State Parameterise simulator $\bar{\psi}^{sim}_{\bar{\phi}}$ with sampled $\bar{\phi}$
\State Reset simulator $\bar{\psi}^{sim}_{\bar{\phi}}$ to a random initial state

\State //Train $RAT$ initial policy using standard $RAT$ training routine (Algo. \ref{alg:rat-train}), but with $\bar{\psi}^{sim}_{\bar{\phi}}$ set as the real-world $\psi_{\phi}$ and $RAT$ policy $\pi_{\theta_{g},\mu_{g}}^{RAT}$ initialised with the policy $\bar{\pi}_{\theta_{g}}^{RAT}$
\State $\bar{\pi}_{\theta_{g}}^{RAT} \gets Train\_RAT\_Policy(\psi^{sim}_{\theta}, \bar{\psi}^{sim}_{\bar{\phi}}, \bar{\pi}_{\theta_{g}}^{RAT})$
\EndFor

\end{algorithmic} 
\end{algorithm}

\subsection{Zero-shot Transfer in Adjacent Environments\label{subsec:drifting-environment}}

Once \emph{RAT} initial policy is learnt in simulation, we examine whether
it can be used in adjacent environments where the reality-gap stays
consistent (i.e., zero-shot transfer). For this purpose, we consider
a modellable latent parameter setting $\hat{\theta}$ that is $\epsilon-$close
($\epsilon<\epsilon{}_{max}$) to the original ground truth parameters
$\theta_{g}$, while the reality-gap between the simulator and the
real-world $\mu_{g}$ remains consistent for both environments. For
such a real-world environment $\psi_{\hat{\phi}}$ (parameterised
by $\hat{\phi}=[\hat{\theta},\mu_{g}]$), we apply \emph{RAT} initial 
policy to transform the UPN policy conditioned by $\hat{\theta}$
to get the corrected real-world action $a_{t}^{R}$.

\begin{align}
\forall_{\hat{\theta}}, & |\hat{\theta}-\theta_{g}|\le\epsilon_{max}\label{eq:drift-env}\\
a_{t}\gets & \pi_{\hat{\theta}}^{S}(s_{t})\nonumber \\
\Delta a_{t}\gets & \pi_{\theta_{g},\mu_{g}}^{RAT}(s_{t}^{RAT}=[s_{t},a_{t}])\nonumber \\
a_{t}^{R}\gets & a_{t}+\Delta a_{t}\nonumber 
\end{align}

The workflow of using \emph{RAT} policy in adjacent environments is
given in Algorithm \ref{alg:zero-shot}.

\begin{algorithm}[ht] 
\caption{Zero-shot sim-to-real policy transfer with RAT policy and UPN} 
\label{alg:zero-shot} \hspace*{\algorithmicindent}
\textbf{Input:} \begin{algorithmic}[1] 
\State $\psi_{\phi}$ - real-world, $\psi^{sim}_{\theta}$ - a parameterisable simulator
\State $\pi^{S}$ - a trained UPN for the task
\State $\theta_{g}$ - ground-truth modellable parameters
\State $\hat{\theta}$ - the adjacent environment parameters
\State $\epsilon_{max}$ - the maximum allowed distance from the ground truth parameters

\State \textbf{Output:} $G^{*}$ - The zero-shot performance of $RAT$ in the adjacent environment \\

\State //Train the robust $RAT$ initial policy
\State $\bar{\pi}^{RAT} \gets Train\_RAT\_Initial\_Policy$($\pi_{\theta_{g}}^{S}$, $\psi^{sim}_{\theta_{g}}$) Algo. \ref{alg:rat-prior}

\State //initialise $RAT$ policy with $\bar{\pi}^{RAT}$
\State $\pi_{\theta_{g},\mu_{g}}^{RAT} \gets \bar{\pi}^{RAT}$

\State //apply $RAT$ on adjacent environments
\If {$|\hat{\theta}-\theta_{g}| \le \epsilon_{max}$}
\State $a_{t} \gets \pi_{\hat{\theta}}^{S}(s_{t})$
\State $\Delta a_{t} \gets \pi_{\theta_{g},\mu_{g}}^{RAT}(s_{t}^{RAT}=[s_{t},a_{t}])$
\State $G^{*} \gets$ Evaluate $a_{t}+\Delta a_{t}$ on adjacent environment (Eq. \ref{eq:drift-env})
\EndIf

\end{algorithmic} 
\end{algorithm}

\subsection{An Alternative \emph{Supervised RAT} Model \label{subsec:supervised-rat}}

The overall objective of \emph{RAT} is to learn a mapping from the simulated action 
to its corresponding real-world action as given in Eq. \ref{eq:rat-correction}. For this purpose, 
instead of using RL to learn a corrective action, we examine an alternative $supervised\:RAT$ model 
where we learn a function $f: (s^{R}_t, s^{R}_{t+1}) \rightarrow a^R_t$ using supervised learning, 
where $s^{R}_t, s^{R}_{t+1} \in \mathcal{S}$ are two subsequent states in the real-world following the
execution of action $a^R_t \in \mathcal{A}$ in the real-world. Then such a function $f$ can be used to map
two states $s_t, s_{t+1} \in \mathcal{S}$ fetched from the simulator after executing UPN action $a_t \in \mathcal{A}$ 
to map the same real-world states to their corresponding real-world action $a^R_t$, $f(s_t, s_{t+1}) \rightarrow a^R_t$.

\emph{Supervised RAT} has a relatively simpler formulation than \emph{RAT}, which also makes it simpler to learn without 
requiring to calibrate RL specific hyperparameters. However, one drawback with \emph{Supervised RAT} is that 
in the presence of large reality-gaps, supervised learning may not
have seen simulation provided states (i.e., the ideal trajectory to follow) and thus not be able to learn the corresponding
real-world action. 

To apply $supervised\:RAT$ model as a zero-shot learner, we follow the same hypothetical real-world setup
discussed in Section \ref{subsec:rat-prior} using the transition function Eq. \ref{eq:rat-prior}. Then it is
evaluated in adjacent environments as discussed in Section \ref{subsec:drifting-environment} by fetching $a^R_t$
directly from $f$ (Eq. \ref{eq:drift-env}).

\section{Experiments}
\subsection{Experimental Setup}

To empirically evaluate our approach, we adopt a set of six continuous
control MuJoCo tasks -- Ant, Half-Cheetah, Hopper, Humanoid, Swimmer
and Walker2D ( \emph{-v2}), implemented with MuJoCo physics simulator
\cite{todorov2012mujoco}. For all tasks, we consider five latent parameters (mass of three body
parts, friction and restitution) that determine the environment configuration,
for which we fetch policies using pretrained UPNs.

To emulate simulated and real-world environments,
for the purpose of this study, two instances of the same MuJoCo simulator
are used, with the real-world environment instance differing from
the simulated instance by virtue of a reality gap. To generate a significant
reality-gap, we have used additional mass and friction for real-world
body-parts of agents ($+400\%$) similar to the approach taken by
\cite{karnan2020reinforced}. Such changes in latent parameters can
be assumed to emulate factors such as air-resistance, muddy soil or
rolling friction which may vary depending on the environment condition.

\subsection{Training \emph{RAT} Policy with UPN}

We use an existing implementation of UPN\footnote{https://github.com/VincentYu68/policy\_transfer}
to store and fetch pretrained policies for MuJoCo tasks. To learn
policies, the UPN is trained for $2\times10^{8}$ steps in the simulator
using Proximal Policy Optimization (PPO) algorithm \cite{DBLP:journals/corr/SchulmanWDRK17},
uniformly sampling latent parameters to learn policies conditioned
on a given set of parameters. To improve the policy at the ground
truth latent parameters, we train another $5\times10^{6}$ steps conditioned
on the ground truth parameters for all tasks except for Walker2D where
$6\times10^{6}$ are used. These values were selected by examining
the simulated task policy performance at each $1\times10^{6}$ steps,
where we stop fine tuning the training if the improvement is less
than $5\%$, which we consider as a criterion to determine that the
simulated task policy has reached convergence.

With the trained UPN, we train both robust \emph{RAT} initial policy 
and \emph{Supervised RAT} using $1\times10^{6}$
simulated steps by uniformly sampling normalised mass and friction
values in the range of $[0,3)$ and $[0,1)$, respectively, from which
the reality-gap is determined by subtracting the normalised ground
truth parameter value. Here, we use a relatively substantial mass
range because of our use of a high mass based reality gap. In reality,
the sampled value range could be determined based on the domain knowledge
of the reality-gap (e.g., wind speed, clay soil stickiness, etc.).
Specifically, for each episode of a task we sample a new reality-gap
value and add it to the original ground truth latent value to generate
the new value, which is used to govern the transition function. Table
\ref{tab:rat-perf-4-0} shows the average step-wise cumulative reward
of \emph{RAT} initialised with the trained initial policy when used in the
real-world without any additional real-world training (i.e., zero-shot
transfer performance). We also present the performance of 
the alternative \emph{Supervised RAT} model trained to directly map simulation
trajectory to real-world actions that would yield a similar trajectory in the real-world.

We use the following UPN-based baselines to
compare \emph{RAT} with.
\begin{itemize}
\item Transfer policy: The simulated policy from UPN conditioned by the
true latent parameter values is transferred to the real-world without
any additional training in the real-world. We evaluate the policy
performance using $100$ episodes in the real-world. 
\item Domain Randomised (DR) policy: $10$ latent parameters $\pm5\%$ from
the ground truth are sampled, for each of which the UPN is conditioned
to fetch the simulated policy and averaged to obtain a final policy.
This generalised DR policy is then applied to the real-world and evaluated
similar to the transfer policy. 
\end{itemize}
To compensate for the $1\times10^{6}$ simulation steps required for
training the \emph{RAT} initial policy, both transfer and DR policies are trained
for an equal amount of additional steps in the simulator.

\begin{table*}
\begin{centering}
\begin{tabular}{|c|c|c|c|c|}
\hline 
 & Transfer & DR & \emph{Supervised RAT} & \emph{RAT}\tabularnewline
\hline 
\hline 
Ant & -0.0363 $\pm$ 0.0021  & -0.0372 $\pm$ 0.0019 & \textbf{0.6374 $\pm$ 0.0015} & 0.228 $\pm$ 0.0016\tabularnewline
\hline 
Half-Cheetah & 0.8607 $\pm$ 0.0049 & 0.8628 $\pm$ 0.0057 & 0.8752 $\pm$	0.0001 & \textbf{0.8817} $\pm$ \textbf{0.0011}\tabularnewline
\hline 
Hopper & 1.6402 $\pm$ 0.0168 & 1.6531 $\pm$ 0.0169 & 1.6620 $\pm$ 0.0193 & \textbf{1.7263} $\pm$ \textbf{0.0221}\tabularnewline
\hline 
Humanoid & 4.9787 $\pm$ 0.0101 & 4.9841 $\pm$ 0.0114 & \textbf{5.2504 $\pm$ 0.0004} & 5.0845 $\pm$ 0.0124\tabularnewline
\hline 
Swimmer & 0.0401 $\pm$ 0.0006 & 0.0399 $\pm$ 0.0006 & 0.0208 $\pm$ 0.0007 & \textbf{0.0436} $\pm$ \textbf{0.0005}\tabularnewline
\hline 
Walker2D & 0.8282 $\pm$ 0.0002 & \textbf{0.8348} $\pm$ \textbf{0.0028} & 0.4768	$\pm$ 0.013 & 0.8077 $\pm$ 0.008\tabularnewline
\hline 
\end{tabular}
\par\end{centering}
\caption[The performance of \emph{RAT} initial policy against baselines in six
MuJoCo environments]{\label{tab:rat-perf-4-0}The zero-shot transfer performance (average step-wise cumulative reward) of \emph{RAT}
policy and \emph{Supervised RAT} against baselines in six MuJoCo environments. Performance
is evaluated on 100 episodes with a horizon of 500 steps ($\pm$ std. error).}
\end{table*}

\subsection{Performance in Adjacent Environments}

We use the \emph{RAT} policy hyperparameter tuned to a real-world
setting to examine its applicability when the latent parameters deviate
from the original ground truth values. For this purpose, we uniformly
sample $100$ latent parameters, deviating $\pm5\%$ from the ground
truth latent parameter values and apply the \emph{RAT} policy, without
any additional training, to transform the UPN policy conditioned on
each of the sampled latent values to ground-truth conditions set to
the same sampled latent values. Table \ref{tab:rat-perf-4-2} shows
the average step-wise cumulative reward of \emph{RAT} policy on these
adjacent latent parameter configurations. 

\begin{table*}
\begin{centering}
\begin{tabular}{|c|c|c|c|c|}
\hline 
 & Transfer & DR  & \emph{Supervised RAT} & \emph{RAT}\tabularnewline
\hline 
\hline 
Ant & -0.0372 $\pm$ 0.0004 & -0.0372 $\pm$ 0.0004 & \textbf{0.6346 $\pm$ 0.0009} & 0.2288 $\pm$ 0.0011\tabularnewline
\hline 
Half-Cheetah & 0.8541 $\pm$ 0.0016 & 0.857 $\pm$ 0.0015 & \textbf{0.8752 $\pm$ 0.0000} & \textbf{0.8749} $\pm$ \textbf{0.001}\tabularnewline
\hline 
Hopper & 1.6532 $\pm$ 0.0004 & 1.6519 $\pm$ 0.0004 & 1.6614	$\pm$ 0.0004 & \textbf{1.7548} $\pm$ \textbf{0.0006}\tabularnewline
\hline 
Humanoid & 4.9964 $\pm$ 0.0019 & 4.9971 $\pm$ 0.0021 & \textbf{5.2443 $\pm$ 0.0008} & 5.0791 $\pm$ 0.0021\tabularnewline
\hline 
Swimmer & 0.0394 $\pm$ 0.0002 & 0.0391 $\pm$ 0.0002 & 0.0209 $\pm$ 0.0003 & \textbf{0.0416} $\pm$ \textbf{0.0002}\tabularnewline
\hline 
Walker2D & \textbf{0.919 }$\pm$ \textbf{0.0177} & 0.8459 $\pm$ 0.005 & 0.4716 $\pm$ 0.003 & 0.8056 $\pm$ 0.003\tabularnewline
\hline 
\end{tabular}
\par\end{centering}
\caption[The performance of \emph{RAT} policy, applied to $100$ uniformly
sampled different latent parameter conditions $\pm\%5$ deviated from
the ground truth values]{\label{tab:rat-perf-4-2}The zero-shot adaptation performance (average step-wise cumulative reward) of \emph{RAT}
policy and \emph{Supervised RAT} for 
$100$ uniformly sampled latent parameter configurations deviating
$\pm\%5$ from the original ground truth values. Performance is evaluated on 100 episodes with
a horizon of 500 steps ($\pm$ std. error).}
\end{table*}

These results illustrate that \emph{RAT} policy is robust to a degree
of deviation from the original ground truth values. While \emph{Supervised RAT} 
performs better in environments such as Ant and Humanoid, it
shows collapses in performance compared to the baselines in Swimmer and Walker2D, 
likely because of not having seen target simulation trajectories in the real-world training data.
In contrast, \emph{RAT} policy shows stable and improved performance compared to the baselines,
perhaps due to better exploration provided by RL.
In the Walker2D environment, \emph{RAT} performs marginally poorer than the baselines
which could be due to the unavailability of a policy in the real-world
to imitate the simulated policy. However, in general, these results
demonstrate that when a simulated task-specific policy has converged
in the simulator, training \emph{RAT} with a small number of additional
simulated experience can be beneficial as a sim2real zero-shot transfer
policy when compared to the baselines.

\section{Conclusion}
In this study, we have proposed a novel policy transformation approach
for adaptable sim2real transfer on unseen environments. The proposed
approach, \emph{Reverse Action Transformation (RAT),} learns a policy
to correct real-world policies such that the resulting trajectory
closely follows the trajectory corresponding to a trained simulated
policy. This learnt correction policy is then deployed in unseen real-world
environments in conjunction with the corresponding simulated policies
from a UPN to achieve zero-shot adaptation. We have empirically evaluated
our approach in six MuJoCo tasks, and demonstrated its superior performance
as a zero-shot learner compared to the relevant baselines.

\pagebreak
\clearpage
\appendix

\bibliographystyle{splncs04}
\bibliography{physics}

\end{document}


\maketitle

\section{Appendix\label{sec:supp}}
\subsection{Source Code}

The source code for this study is shared in the \emph{code} folder. Follow the instructions in \emph{README} to install and run the project.

\subsection{Implementation Details \label{subsec:implementation-details}}

To learn \emph{RAT} we used the same PPO implementation used for UPN,
which uses Generalised Advantage Estimate (GAE) \cite{DBLP:journals/corr/SchulmanMLJA15}
in its target value calculation and Adam \cite{kingma2015adam} as
an optimiser. When learning \emph{RAT }we optimised PPO specific hyperparameters,
discount factor (\emph{$\gamma$}), Clip threshold and Entropy Coefficient,
optimiser's step size ($\alpha$), and GAE smoothing parameter ($\lambda$).
To tune them, we used Bayesian Optimisation (BO) \cite{frazier2018tutorial}
with $40$ iterations for each task, using the original real-world
performance as the objective. In addition to these hyperparameters,
we considered resetting simulation trajectory to match the real-world
at each state (i.e., matching each transition trajectory segments
separately) as another hyperparameter, which has been shown to improve
the performance of certain tasks. When trajectories were not reset,
\emph{RAT }was trained to correct the entire episodic trajectory instead
of separate transition segments. To learn \emph{RAT }policy, we used
$5$ or $10$ hidden layers of 128 neurons depending on the task's
complexity and reset option, with a batch size of $4000$ for all
tasks.

For all tasks, a single original ground truth latent parameter set
was generated uniformly in $[0,1)$ range, {[}$0.5488135$, $0.71518937$,
$0.60276338$, $0.54488318$, $0.4236548${]}, representing parameters
friction, mass of torso and two other body parts, and coefficient
of restitution, respectively.

\subsection{Extended Results}

\subsubsection{Performance Limits of \emph{RAT }in Adjacent Environments}

To examine the limits of \emph{RAT's} performance in\emph{ }deviating
latent parameter settings, we have evaluated \emph{RAT} in settings
that are within $\pm10\%$ and $\pm20\%$ maximum deviation from the
ground truth latent parameter, of which results are presented in tables
\ref{tab:rat-perf-4-3} and \ref{tab:rat-perf-4-4}, respectively.
We observe that \emph{RAT} is fairly robust to up to $\pm10\%$ latent
parameter deviation from the ground truth, but loses its advantage
when reaching $\pm20\%$ deviation.

\begin{table*}
\begin{centering}
\subfloat[\label{tab:rat-perf-4-3} $\pm\%10$ maximum deviation from the ground
truth]{
\begin{centering}
\begin{tabular}{|c|c|c|c|}
\hline 
 & Transfer & DR & \emph{RAT}\tabularnewline
\hline 
\hline 
Ant & -0.0373$\pm$ 0.0011 & -0.0323$\pm$ 0.0004 & \textbf{0.2277 $\pm$ 0.0021}\tabularnewline
\hline 
Half-Cheetah & 0.8454 $\pm$ 0.0037 & \textbf{0.8610 $\pm$ 0.0014} & \textbf{0.8613 $\pm$ 0.0021}\tabularnewline
\hline 
Hopper & 1.6527 $\pm$ 0.0009 & 1.6506 $\pm$ 0.0004 & \textbf{1.7511 $\pm$ 0.0008}\tabularnewline
\hline 
Humanoid & 4.9949 $\pm$ 0.0034 & 4.9942$\pm$ 0.0019 & \textbf{5.0771 $\pm$ 0.0036}\tabularnewline
\hline 
Swimmer & 0.0397 $\pm$ 0.0003 & 0.0382 $\pm$ 0.0002 & \textbf{0.0406 $\pm$ 0.0004}\tabularnewline
\hline 
Walker2D & \textbf{1.0022 $\pm$ 0.0274} & 0.8741 $\pm$ 0.0094 & 0.8538 $\pm$ 0.0126\tabularnewline
\hline 
\end{tabular}
\par\end{centering}

}\\\subfloat[\label{tab:rat-perf-4-4}$\pm\%20$ maximum deviation from the ground
truth]{\begin{centering}
\begin{tabular}{|c|c|c|c|}
\hline 
 & Transfer & DR & \emph{RAT}\tabularnewline
\hline 
\hline 
Ant & -0.0229\textbf{ $\pm$ }0.0039 & -0.0169 \textbf{$\pm$ }0.0004 & \textbf{0.2258 $\pm$ 0}.\textbf{0040}\tabularnewline
\hline 
Half-Cheetah & 0.8340 \textbf{$\pm$ }0.0065 & \textbf{0.8671 $\pm$ 0.0014} & 0.8342 \textbf{$\pm$} 0.0039\tabularnewline
\hline 
Hopper & 1.6532 \textbf{$\pm$ }0.0016 & 1.6489 \textbf{$\pm$ }0.0004 & \textbf{1.7471 $\pm$ 0.0014}\tabularnewline
\hline 
Humanoid & 4.9894 \textbf{$\pm$ }0.0064 & 4.9862 \textbf{$\pm$ }0.0019 & \textbf{5.0676 $\pm$ 0.0070}\tabularnewline
\hline 
Swimmer & \textbf{0.0413 $\pm$ 0.0007} & 0.0381 \textbf{$\pm$ }0.0001 & 0.0368 \textbf{$\pm$ }0.0010\tabularnewline
\hline 
Walker2D & \textbf{1.0079 $\pm$ 0.0303} & \textbf{0.9932 $\pm$ 0.0199} & 0.9309 \textbf{$\pm$} 0.0238\tabularnewline
\hline 
\end{tabular}
\par\end{centering}

}\caption[The performance of \emph{RAT} policy, applied to $100$ uniformly
sampled different latent parameter conditions $\pm\%10$ and $\pm\%20$
deviated from the ground truth values]{The zero-shot transfer performance (average
step-wise cumulative reward) of \emph{RAT} policy applied to $100$ uniformly
sampled latent parameter configurations deviating $\pm\%10$ and $\pm\%20$
from the original ground truth values. Performance is evaluated on 100 episodes with a horizon
of 500 steps ($\pm$ std. error).}
\end{centering}
\end{table*}


\subsection{Machine Details}

\begin{itemize}
\item CPU: AMD EPYC 7742 Processor
\item CPU cores: 128
\item CPU speed: ~3400 MHz
\item Address sizes: 43 bits physical, 48 bits virtual
\item cache size: 512 KB
\item Memory: 1007.5 GB

\end{itemize}

\bibliographystyle{named}
\bibliography{physics}